\documentclass[10pt,twocolumn,letterpaper]{article}

\usepackage{cvpr}
\usepackage{times}
\usepackage{epsfig}
\usepackage{graphicx}
\usepackage{amsmath}
\usepackage{amssymb}

\cvprfinalcopy 

\def\cvprPaperID{}

\begin{document}

\title{Global Pose Estimation with an Attention-based Recurrent Network 
}

\author{Emilio Parisotto$^{*,1,2}$, Devendra Singh Chaplot$^{*,1,2}$, Jian Zhang$^1$, Ruslan Salakhutdinov$^{1,2}$ \\
  $^*$Equal Contribution.\\
  $^1$Apple Inc., 1 Infinite Loop, Cupertino, CA 95014, USA.\\
  $^2$Carnegie Mellon University, 5000 Forbes Ave, Pittsburgh, PA 15213, USA.\\
  {\tt\small \{eparisot,dchaplot\}@cs.cmu.edu \qquad \{jianz,rsalakhutdinov\}@apple.com}
 }

\newcommand{\fix}{\marginpar{FIX}}
\newcommand{\new}{\marginpar{NEW}}

\maketitle

\newcommand\methodname{Neural Graph Optimizer}

\begin{abstract}

%

The ability for an agent to localize itself within an environment is crucial for many real-world applications. 
For unknown environments, Simultaneous Localization and Mapping (SLAM) enables incremental and concurrent building of and localizing within a map. 
We present a new, differentiable architecture, \methodname, progressing towards a complete neural network solution for SLAM by designing a system composed of a local pose estimation model, a novel pose selection module, and a novel graph optimization process. 
The entire architecture is trained in an end-to-end fashion, enabling the network to automatically learn domain-specific features relevant to the visual odometry and avoid the involved process of feature engineering.
We demonstrate the effectiveness of our system on a simulated 2D maze and the 3D ViZ-Doom environment.
\end{abstract}

\begin{figure}[t!]
	  \centering
   \includegraphics[width=\linewidth,height=\textheight,keepaspectratio]{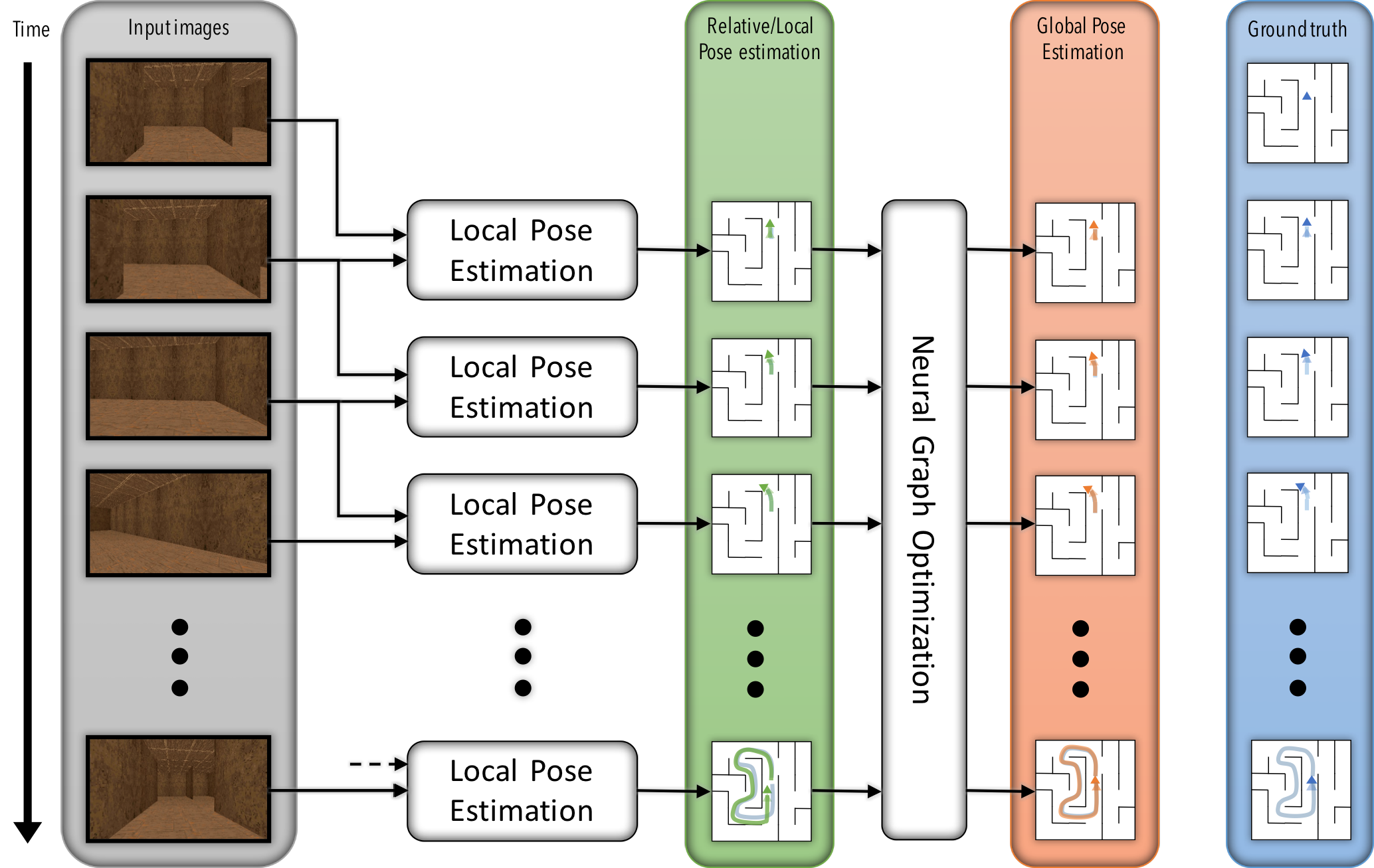}
	\caption{\small{Components of the proposed model along with sample input, output and ground truth. The Local Pose Estimation model predicts the relative pose change between consecutive observations and Neural Graph Optimization model jointly optimizes the predictions of the Local Pose Estimation model to predict global pose changes. The local pose estimates, global pose estimates, and ground truth trajectory are shown in green, orange and blue, respectively.}}
  \label{fig:env}
\end{figure}

\section{Introduction}

The ability for an agent to localize itself within an environment is a crucial prerequisite for many real-world applications, such as household robots \cite{thrun2005probabilistic}, autonomous drones \cite{forster2014svo}, augmented and virtual reality applications, and video game AI \cite{parisotto2017neural}. In most cases, the main challenge for an agent localizing itself is that, the agent is not provided with a map of the environment and therefore the agent must simultaneously map the environment and localize itself within the incomplete map it has produced. A wide variety of algorithms to solve this Simultaneous Localization and Mapping (SLAM) task have been developed over a long history \cite{thrun2005probabilistic,cadena2016past}, with modern methods achieving impressive accuracy and real-time performance \cite{mourikis2007multi,kummerle2011g,mur2015orb,engel2014lsd}. These methods still have several shortcomings, owing mainly to the hand-engineered features, dense matching, and heuristics used in the design of these algorithms. For example, most methods are brittle in certain scenarios, such as varying lighting conditions (e.g.\ changing time of day), different weather conditions or seasons \cite{sattler2017benchmarking}, repetitive structures, textureless objects, extremely large viewpoint changes, dynamic elements within the environment, and faulty sensor calibration \cite{cadena2016past}. Because these situations are common in real-world scenarios, robust applications of those systems are difficult.

In this paper, we develop a method which can be made more robust to the common situations where previous SLAM algorithms typically degrade. 
To do this, 
we formulate a novel neural network architecture called ``\methodname''. 
\methodname~consists of differentiable analogues of the common types of subsystems used in modern SLAM algorithms, such as a local pose estimation model, a pose selection module (key frame selection, essential graph), and a graph optimization process. 
Because each component in the system is differentiable, the entire architecture can be trained in an end-to-end fashion, 
 enabling the network to learn invariances to the types of 
scenarios observed during training. 

To demonstrate the ability of our method to learn pose estimation, we use trajectories sampled from several simulated environments. The first environment is a 2D maze where the agent has a single-pixel row-scan as input. We then scale the model up to 3D mazes based on the ViZDoom environment \cite{kempka2016vizdoom}, where the agent receives an image of the  first-person view of the world as input. 

\section{Related Work}

SLAM is a process in which an agent needs to localize itself in an unknown environment and build a map of this environment at the same time, with uncertainties in both its motions and observations. SLAM has evolved from filter-based to graph-based (optimization-based) approaches. Some EKF-based systems have demonstrated state-of-the-art performance, such as the Multi-State Constraint Kalman Filter  \cite{mourikis2007multi}, the VIN \cite{kottas2013consistency}, and the system of Hesch et al. \cite{hesch2014camera}. Those methods, even though efficient, heavily depend on linearization and Gaussian assumptions, and thus under-perform their optimization-based counterparts, such as OK-VIS \cite{leutenegger2015keyframe}, ORB-SLAM \cite{mur2015orb}, and LSD-SLAM \cite{engel2014lsd}. 

Graph-based SLAM typically includes two main components: the front-end and the back-end. The front-end extracts relevant information (e.g.\ salient features) from the sensor data and associates each measurement to a specific map feature, while the back-end performs graph optimization on a graph of abstracted data produced by the front-end.

Graph-based SLAM can be categorized either as feature-based or direct methods depending on the type of front-end. Feature-based methods rely on local features (e.g.\ SIFT, SURF, FAST, ORB, etc.) for pose estimation. For example, ORB-SLAM \cite{mur2015orb} performs data association and camera relocalization with ORB features and DBoW2 \cite{galvez2012bags}. RANSAC~\cite{fischler1987random} is commonly used for geometric verification and outlier rejection, and there are also prioritized feature matching approaches \cite{sattler2016efficient}. However, hand-engineered feature detector and descriptors are not robust to motion blur, illumination changes, or strong viewpoint changes, any of which can cause localization to fail. 

To avoid some of the aforementioned drawbacks of feature-based approaches, direct methods, such as LSD-SLAM \cite{engel2014lsd}, utilize extensive photometric information from the images to determine the pose, by minimizing the photometric error between corresponding pixels. This approach is in contrast to feature-based methods, which minimize the reprojection error. However, such methods are usually not applicable to wide baseline settings \cite{cadena2016past} during large viewpoint changes. Recent work in \cite{forster2014svo} \cite{forster2017svo} combines feature and direct methods by minimizing the photometric error of features lying on intensity corners and edges. Some methods focus on dense recontruction of the scene, for instance \cite{whelan2016elasticfusion} builds dense globally consistent surfel-based maps of room scale environments explored using an RGB-D camera, without pose graph optimisation, while KinectFusion \cite{newcombe2011kinectfusion} obtains depth measurements directly using active sensors and fuse them over time to recover high-quality surface maps. These approaches still suffer from strict calibration and synchronization requirements, and the data association modules require extensive parameter tuning in order to work correctly for a given scenario. 

In light of the limitations of feature-based and direct approaches, deep networks are proposed to learn suitable feature representations that are robust against motion blur, occlusions, dynamic scenes, illumination, texture, and viewpoint changes. They have been successfully applied to several related multiview vision problems, including learning optical flow \cite{dosovitskiy2015flownet}, depth \cite{liu2015deep}, homography between frame pairs \cite{detone2016deep}, and localization \cite{chaplot2018active} and re-localization problems. 

Recent work includes re-formulating the localization problem as a classification task \cite{weyand2016planet}, a regression task \cite{kendall2015posenet,hazirbasimage2017}, end-to-end trainable filtering \cite{haarnoja2016backprop}, and differentiable RANSAC \cite{brachmann2016dsac}. More specifically, PlaNet \cite{weyand2016planet} formulates localization as a classification problem, predicting the corresponding tile from a set of tiles subdividing Earth surface for a given image, thus providing the approximate position from which a photo was taken. PoseNet \cite{kendall2015posenet} formulates 6-DoF pose estimation as a regression problem. One drawback of the PoseNet approach is its relative inaccuracy, compared to state-of-the-art SIFT methods. Similarly, \cite{melekhov2017relative} fine-tunes a pretrained classification network to estimate the relative pose between two cameras. To improve its performance, \cite{hazirbasimage2017} added Long-Short Term Memory (LSTM) units to the fully-connected layers output, to perform structured dimensionality reduction, choosing the most useful feature correlations for the task of pose estimation. From a different angle, DSAC \cite{brachmann2016dsac} proposes a differentiable RANSAC so that a matching function that optimizes pose quality can be learned. These approaches are not robust to repeated structure or similar looking scenes, as they ignore the sequential and graphical nature of the problem. Addressing this limitation, work in \cite{clark2017vinet} fused additional sequential inertial measurement with visual odometry.
SemanticFusion \cite{mccormac2017semanticfusion} combines convolutional neural networks (CNNs) and a dense ElasticFusion \cite{whelan2016elasticfusion}. However, classic feature-based methods still outperform CNN-based methods published to date in terms of accuracies. 

Recently, there has been an increasing interest in combining navigation and plannning in an end-to-end deep reinforcement learning (DRL) framework. The efforts to date can be divided into two categories depending on the presence of external memory in the architecture or not.
Target-driven visual navigation takes a visual observation and an image of the target \cite{zhu2017target} or range findings \cite{tai2017virtual} as input, and plans goal seeking actions in a 3D indoor simulated environment as the output.

In simulated environments, \cite{mirowski2016learning} uses stacked LSTM in a goal-driven RL problem with auxilary tasks of depth prediction and loop-closure classification, while \cite{zhang2016deep} added successor features to ease transfer from previously mastered navigation tasks to new ones. Work in \cite{bhatti2016playing} augmented DRL with Faster-RCNN for object detection and SLAM (ORB-SLAM2) for pose estimation; observing images and depth from VizDoom, they built semantic maps with 3D reconstruction and bounding boxes as input to a RL policy. 

To deal with the limited memory of standard recurrent architures (such as LSTM) more structured external memories have been developed to take the spatial relations of memories into account. \cite{gupta2017cognitive} assumes known ego-motion and constructs a metric egocentric multi-scale belief map (top-down-view latent representation of free space) of the world with a 2D spatial memory, upon which RL plans a sequence of actions towards goals in the environment with a value iteration network. Neural Map in \cite{parisotto2017neural} is a writable structured 2D external memory map for an agent to learn to navigate within 2D and 3D maze environments. These works all assume precise egomotion and thus perfect localization, a prerequisite that can rarely be met in real-world scenarios. Relaxing this assumption and resembling traditional occupancy grid SLAM, Neural SLAM \cite{zhang2017neural} uses an occupancy-grid-like memory map, assuming only an initial pose is provided, and updates the pose beliefs and grid map using end-to-end DRL.

One of key ingredient for the success of graph-based SLAM is the back-end optimization. The back-end builds the pose graph, in which two pose nodes share an edge if an odometry measurement is available between them, while a landmark and a robot-pose node share an edge if the landmark was observed from the corresponding robot pose. In pose graph optimization, the variables to be estimated are poses sampled along the trajectory of the robot, and each factor imposes a constraint on a pair of poses. Modern SLAM solvers exploit the sparse nature of the underlying factor graph and apply iterative linearization and optimization methods (e.g.\, nonlinear least squares via the Gauss-Newton or Levenberg-Marquardt algorithm). Several such solvers achieve excellent performance, for example, g2o \cite{kummerle2011g}, TSAM \cite{dellaert2012factor}, Ceres, iSAM \cite{kaess2012isam2}, SLAM++ \cite{salas2013slam}, and recently  \cite{bowman2017probabilistic} for optimization with semantic data association. The SLAM back-end offers a natural defense against data association and perceptual aliasing errors from the front-end, where similarly looking scenes, corresponding to distinct locations in the environment, would deceive place recognition. However, they depend heavily on linearization of the sensing and motion models, and require good initial guesses.  Current systems can be easily induced to fail when either the motion of the robot or the environment are too challenging (e.g.\, fast robot dynamics or highly dynamic environments) \cite{cadena2016past}.

In this work we formulate a complete end-to-end trainable solution to the graph-based SLAM problem. 
We present a novel architecture that combines a CNN-based local front-end and an attention-based differentiable back-end. We learn effective features automatically and perform implicit loop closure by designing an additional differentiable Neural Graph Optimizer to perform global optimization over entire pose trajectories and correct errors accumulated by the local estimation model. 

\begin{figure*}
\vspace{-0.1in}
  \includegraphics[width=\linewidth,height=\textheight,keepaspectratio]{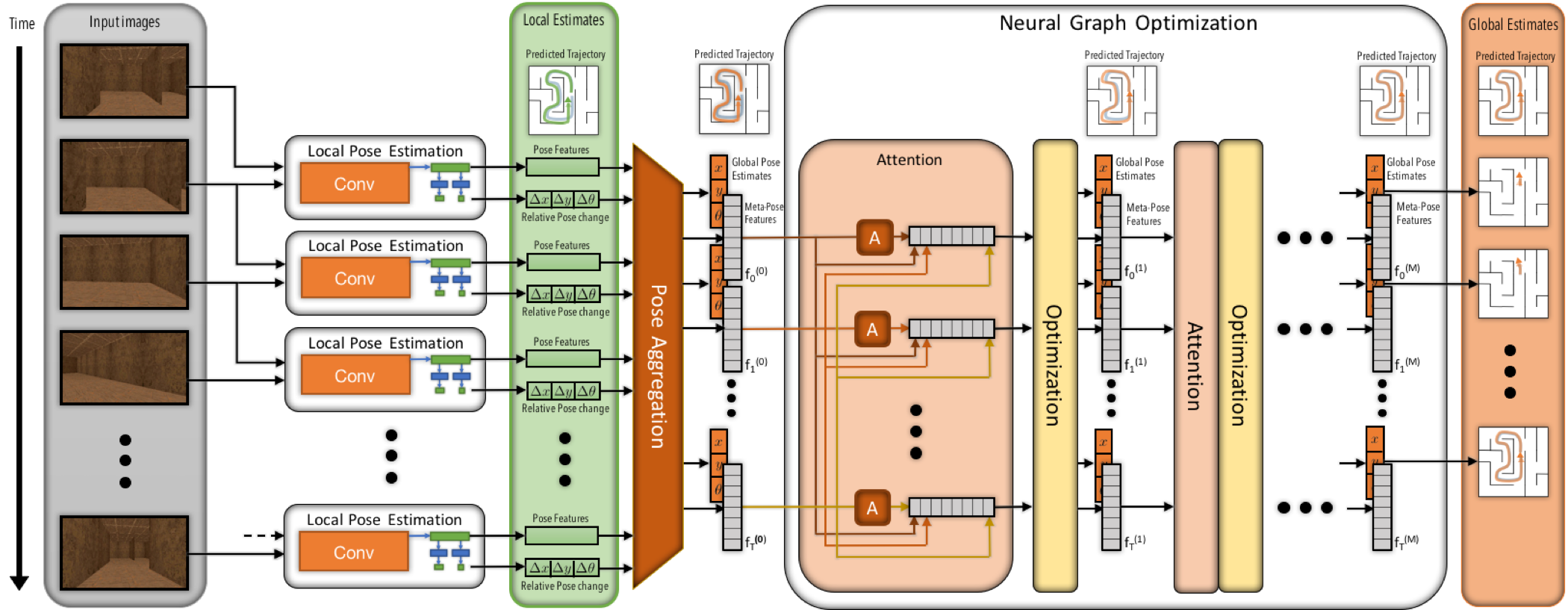}
\caption{\small{The architecture of the proposed model, showing the Local Pose Estimation, the Pose Aggregation, and the Neural Graph Optimization modules.}}
\label{fig:global_arch}
\vspace{-0.1in}
\end{figure*}

\section{Method}

The \methodname~architecture is split into distinct differentiable components. 
Similar to many of the previous methods, we split the process into local adjustments between temporally adjacent frames combined with a global optimization procedure which distributes error over the entire observed trajectory. As will be shown in the experiments, the global graph optimization procedure is critical to removing drift (the accumulation of small errors over long trajectories). The graph optimization procedure does this by learning to do loop closures, recognizing when the agent has revisted the same location, and enforcing a constraint that those poses should be nearly equal. The local model is crucial for providing a good starting point for the global optimization. It does this by estimating relative transformations between two temporally adjacent frames. By accumulating transformations from the start of the trajectory to the end, we can use this model to get the initial pose estimate within the global frame. 

The complete model architecture is shown in Fig.\ \ref{fig:global_arch}. We will describe relative poses as $\Delta P = (\Delta p_1,\hdots,\Delta p_T)$ with the first pose set as the origin, i.e.\ $\Delta p_1$ is the transformation from origin to pose 1, $\Delta p_2$ is the transformation from pose~1 to pose 2, and so on. These relative poses can be transformed into a global frame of reference by accumulating the relative pose changes along the trajectory, i.e.\ $p_1 = \Delta p_1 \bf{I}$, $p_2 = \Delta p_2 \Delta p_1 \bf{I}$, and so on. These global poses will be refered to as $P = (p_1,\hdots,p_T)$. There exists a differential function $r2g = g2r^{-1}$ such that $P = r2g(\Delta P)$ and $\Delta P = g2r(P)$. 
Each component is described in more detail in the next sections.

\subsection{Local Pose Estimation Network}
\begin{figure*}
\includegraphics[width=\linewidth,height=\textheight,keepaspectratio]{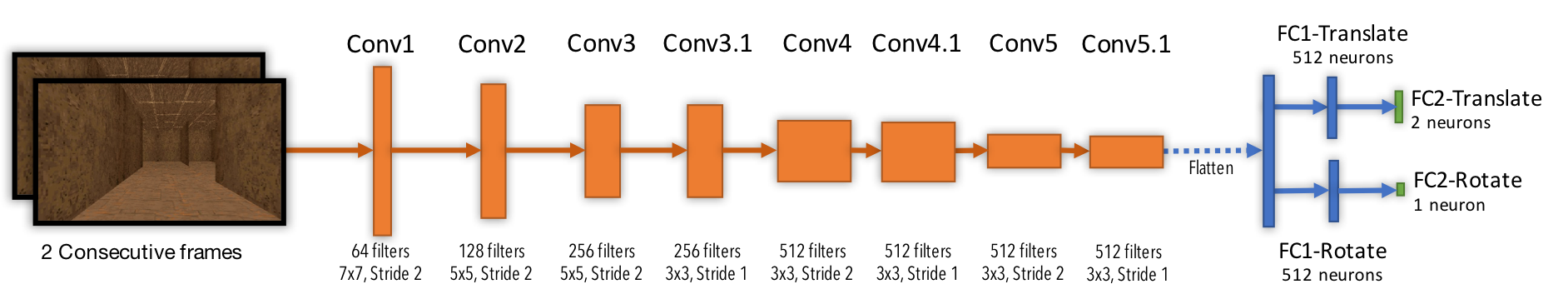}
\caption{\small{The architecture of the Local Pose Estimation network. The architecture of the convolutional layers is adapted from the architecture of the Flownet \cite{dosovitskiy2015flownet}.}}
\label{fig:local_arch_actions}
\vspace{-0.1in}
\end{figure*}

The Local Pose Estimation network learns to predict the relative pose change between two consecutive frames. From two consecutive observations, where each observation is, for example,  an RGB frame, this component predicts the x-coordinate, y-coordinate, and orientation ($\Delta x$, $\Delta y$ and $\Delta \theta$) of the second frame with respect to the first frame. It can also optionally take in side information, such as the action taken by the agent between the two frames. The architecture of the Local Pose Estimation network is shown in Fig.~\ref{fig:local_arch_actions}. Both frames are stacked and passed through a series of convolutional layers. The output of the convolutional layers is flattened and passed to two fully-connected layers that predict the translational and rotational pose change respectively.

Some of the recent work showed that optical flow is useful in predicting frame-to-frame ego-motion \cite{costante2016exploring}. The architecture of the Local Pose Estimation network is inspired by the architecture of Flownet \cite{dosovitskiy2015flownet} which predicts the optical flow between two frames. The convolutional layers in the Local Pose Estimation network are identical to the convolutional layers in Flownet. Prior work on visual odometry and visual inertial odometry has also used the convolutional layer architecture of Flownet \cite{clark2017vinet,wang2017deepvo}.

\subsection{Pose Aggregation}

The next step of the architecture is a Pose Aggregation network which takes in a large number of low-level poses and pose features (up to 2000 for 2D, 1000 for 3D VizDoom environment) and reduces them into a smaller number of more temporally distant ``meta-poses'' and ``meta-pose features''  (around 250 for 2D, 125 for 3D VizDoom). These resulting meta-poses and meta-pose features are then passed to the Neural Graph Optimization procedure. 

For pose feature aggregation, we utilize a deep temporal convolutional network with several alternating layers of (kernel size 3, stride 1, padding 1) dimension-preserving convolutions and (kernel size 2, stride 2, padding~0) dimension-reducing max pooling (where each max pooling operation halves the sequence size). The number of times we halve the sequence length is a hyperparameter. 
Instead of temporal convolutions, we could have utilized recurrent networks, but we decided to focus on convolutions for computational and memory-efficiency reasons. 

In addition to the pose features being aggregated into meta-pose features by the temporal convolution, we also compose all the local pose transformations that were predicted by the Local Pose Estimation model. This composition gives us an initial global pose estimate for each of the meta-poses. The combined meta-features and meta-poses are then passed onto the Neural Graph Optimization layer for the final global pose adjustments,
as shown in Fig.~\ref{fig:global_arch}.

\subsection{Neural Graph Optimization}

The final component of our system is the ``Neural Graph Optimizer''. This submodule aggregates information over the entire pose trajectory with the goal of redistributing error to minimize drift. The Neural Graph Optimizer model is a neural analogue of the global optimization procedures commonly used in traditional state-of-the-art SLAM packages, such the g2o framework~\cite{kummerle2011g}. We define the Neural Graph Optimizer as a recurrent network submodule which takes as input sequential pose features and outputs a refined estimate of these poses. 

In more detail, the Neural Graph Optimizer takes as input some initial $T$ relative pose estimates (i.e. the aggregated output of the local pose estimation network)
$\Delta {\bf P}^{(0)} = \left(\Delta p_1^{(0)},\hdots,\Delta p_T^{(0)}\right)$ 
and produces two outputs for each pose:
\vspace{-0.1in}
\begin{align*}
\nabla {\bf P}^{(1)} &= \left(\nabla p_1^{(1)},\hdots,\nabla p_T^{(1)}\right),
\hspace{0.05in} \textrm{and} \nonumber \\ 
\pmb{\beta}^{(1)} &= \left(\beta_1^{(1)},\hdots,\beta_T^{(1)}\right). 
\end{align*}
New pose estimates are then constructed by performing an iterative update: 
\vspace{-0.1in}
\begin{align*}
\Delta p_i^{(1)} = \Delta p_i^{(0)} + \beta_i^{(1)} \nabla p_i^{(1)}. 
\end{align*}
The Neural Graph Optimizer procedure can then be rerun on the new pose estimates $\Delta {\bf P}^{(1)} = (\Delta p_1^{(1)},\hdots,\Delta p_T^{(1)})$ to produce $\Delta {\bf P}^{(2)} = (\Delta p_1^{(2)},\hdots,\Delta p_T^{(2)})$, and so on. The process is repeated until some pre-specified number of iterations $M$ has taken place. We then transform the refined relative pose estimates into the final global output: ${\bf P}^{(M)} = r2g(\Delta {\bf P}^{(M)})$. 

The specific architecture of the Neural Graph Optimizer is based on two priors that are intuitively useful for pose optimization. The first prior is the notion that poses that are temporally adjacent should have similar outputs, while the second prior is that visually similar but temporally disparate poses should also have similar outputs since this provides a hint that a place has been revisited, thereby potentially enabling a loop closure-like correction of drift. 
We express these priors by using two architectural systems in the Neural Graph Optimizer. The first is a Transformer-like \cite{vaswani2017attention} attention phase where information is propagated over the entire sequence, and the second is a convolutional phase where local temporal information is aggregated.

\subsubsection{Attention Phase}

Suppose there is a meta-pose sequence of $T$ steps, processed by the pose aggregation network into an initial set of features at each time step: ${\bf F}^{(0)} = (f_1^{(0)},\ldots,f_T^{(0)})$.
The attention phase computes, for each pose, a soft-attention operation over the entire trajectory. This attention operation allows each pose to query information over long time spans. The attention phase takes as input the pose feature sequence $(f_1^{(i-1)},\ldots,f_T^{(i-1)})$ and produces for each time step a query vector: $(q_1^{(i-1)},\ldots,q_T^{(i-1)})$ using a fully-connected layer. Then, for each query vector $q_t^{(i-1)}$, a soft-attention operation is carried out to produce an attention vector $a_t^{(i-1)}$ as follows:
\begin{align*}
  C_{tu} &= \langle q_t, f_u \rangle, \\
  \alpha_{tu} &= \frac{C_{tu}}{\sum_{v=1}^T C_{tv}}, \\
  a_t &= \sum_{v=1}^T \alpha_{tu} \odot f_u,
\end{align*}
where the superscripts $(i-1)$ were omitted for clarity of notation.
This produces a sequence of attention vectors $(a_1^{(i-1)},\ldots,a_T^{(i-1)})$, which are passed along with $(f_1^{(i-1)},\ldots,f_T^{(i-1)})$ to the next ``Optimization'' phase.

\subsubsection{Optimization}

The optimization phase aggregates local temporal information by passing the pose features through several temporal convolutions and is responsible for producing the iterative adjustments: $\{\nabla p_1^{(i)},\ldots,\nabla p_T^{(i)}\}$ and $\{\beta_1^{(i)},\ldots,\beta_T^{(i)}\}$.  The optimization phase proceeds as follows: First, the attention and feature vectors are concatenated into a new sequence of features:
\vspace{-0.1in}
\begin{align*}
\left(\begin{bmatrix}
    f_1^{(i-1)} \\
    a_1^{(i-1)} 
  \end{bmatrix},\ldots,\begin{bmatrix} 
    f_T^{(i-1)} \\ 
    a_T^{(i-1)} 
  \end{bmatrix}\right).
\end{align*}
These features are then passed through several layers of 1D convolutions $h_l$ and activations $\sigma_l$:
\begin{align*}
  \begin{bmatrix}
    {\bf F}^{(i)} \\
    \nabla {\bf P}^{(i)} \\
    \pmb{\beta}^{(i)}
  \end{bmatrix} =  \sigma_L\left(h_L\left(...~h_1\left(\begin{bmatrix}
    f_1^{(i-1)} \\
    a_1^{(i-1)}
  \end{bmatrix}...\begin{bmatrix}
        f_T^{(i-1)} \\
        a_T^{(i-1)}
  \end{bmatrix}\right) ... \right)\right)
\end{align*}
to produce the current iteration's adjustments ($\nabla {\bf P}^{(i)}$ and $\pmb{\beta}^{(i)}$) as well as the feature layer for the next iteration of the process (${\bf F}^{(i)}$).

For our experiments, we use 9 layers of convolutions with filter size 3 and ReLU activations. While temporal convolutions have a limited receptive field which provides a hard upper limit on how far they can transmit information across time, we found that in practice they worked better than using a bidirectional LSTMs.

\subsubsection{Induced Attention Graph}

We now provide some intuition on why the attention phase enables higher performance than only using the optimization phase, or running all pose features through bidirectional LSTMs. 
We can see that during the attention phase, some similarity graph $C$ is constructed such that each element $C_{tu}$ is the inner product between the query vector~$q_t$ and the pose feature vector $f_u$. Therefore $C$ represents a similarity matrix between the queries and pose features, and those with very similar features will thus have high information bandwidth through the attention operator because the attention weight $\alpha_{tu}$ will be near $1$ for highly similar query and pose features, and near $0$ otherwise.   
The attention operation is thus inducing a connectivity graph between poses with highly similar features. This therefore resembles a soft, differentiable analogue of the pose graph constructed in SLAM algorithms such as ORB-SLAM \cite{mur2015orb}.

\begin{figure}
  \centering
  \minipage{0.48\linewidth}
    \centering
    \includegraphics[width=\linewidth,height=\textheight,keepaspectratio]{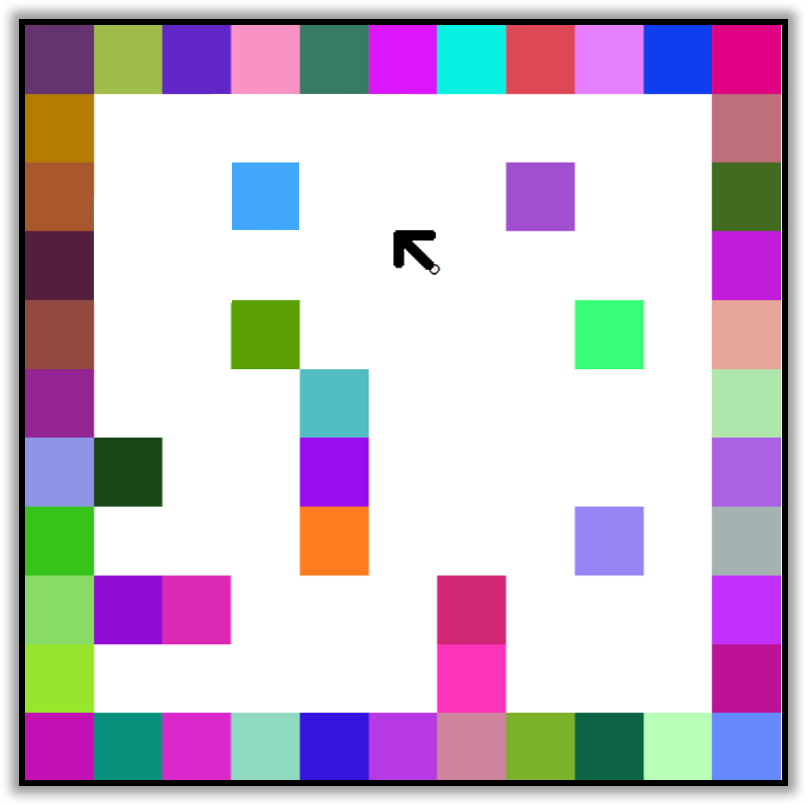}
  \endminipage \hfill
  \minipage{0.48\linewidth}
    \centering
    \includegraphics[width=\linewidth,height=\textheight,keepaspectratio]{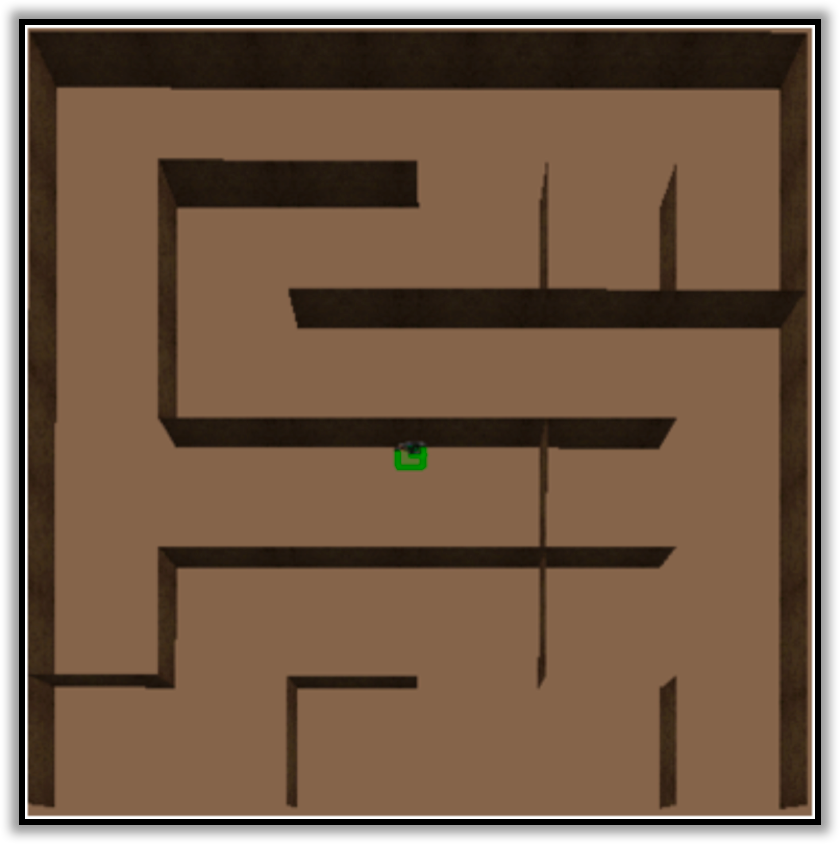}
  \endminipage
\vspace{0.05in}
  \caption{\small{\textbf{Left:} A screenshot of the 2D environment based on Box2D. \textbf{Right:} A bird's eye view of the 3D environment based on the Doom game engine.}}
  \label{fig:env}
\end{figure}

\begin{table}
\small
  \centering
	\begin{tabular}{|c|c|}
		\hline
      \multicolumn{2}{|c|}{\bf{Results on the 2D Environment}} \\ \hline
		{\bf Model} & {\bf RMSE} \\ \hline
		Only Local Estimation                  & 17.80 \\ \hline
		Global Estimation - 1 Attend-Opt iteration  & 10.21  \\
		Global Estimation - 5 Attend-Opt iterations  & 3.16   \\ \hline
	\end{tabular}
\vspace{0.1in}
	\caption{\label{tab:2dngo} Results for different Neural Graph Optimizer architectures and hyperparameters, in terms of test set Global RMSE. We can see that the addition of the global optimization procedure reduces the loss by more than 80\% as compared to solely using the local pose model. }
\end{table}

\section{Experiments}
We use two simulation environments for our experiments, a 2D environment based on Box2D and a 3D environment based on the Doom game engine. To train the system, we pretrained the local pose estimation model and then trained the global optimizer with the local pose model held fixed. This was mainly due to the large sequence lengths we were required to process (on the order of 1000 time steps). This limited the amount of sequences we could process due to the large memory requirements. Training the system in stages enabled us to preprocess the sequence images into a far more memory-efficient compressed representation. 

\begin{figure}
	\centering
	\includegraphics[trim = 7mm 7mm 7mm 7mm,clip, width=0.49\linewidth]{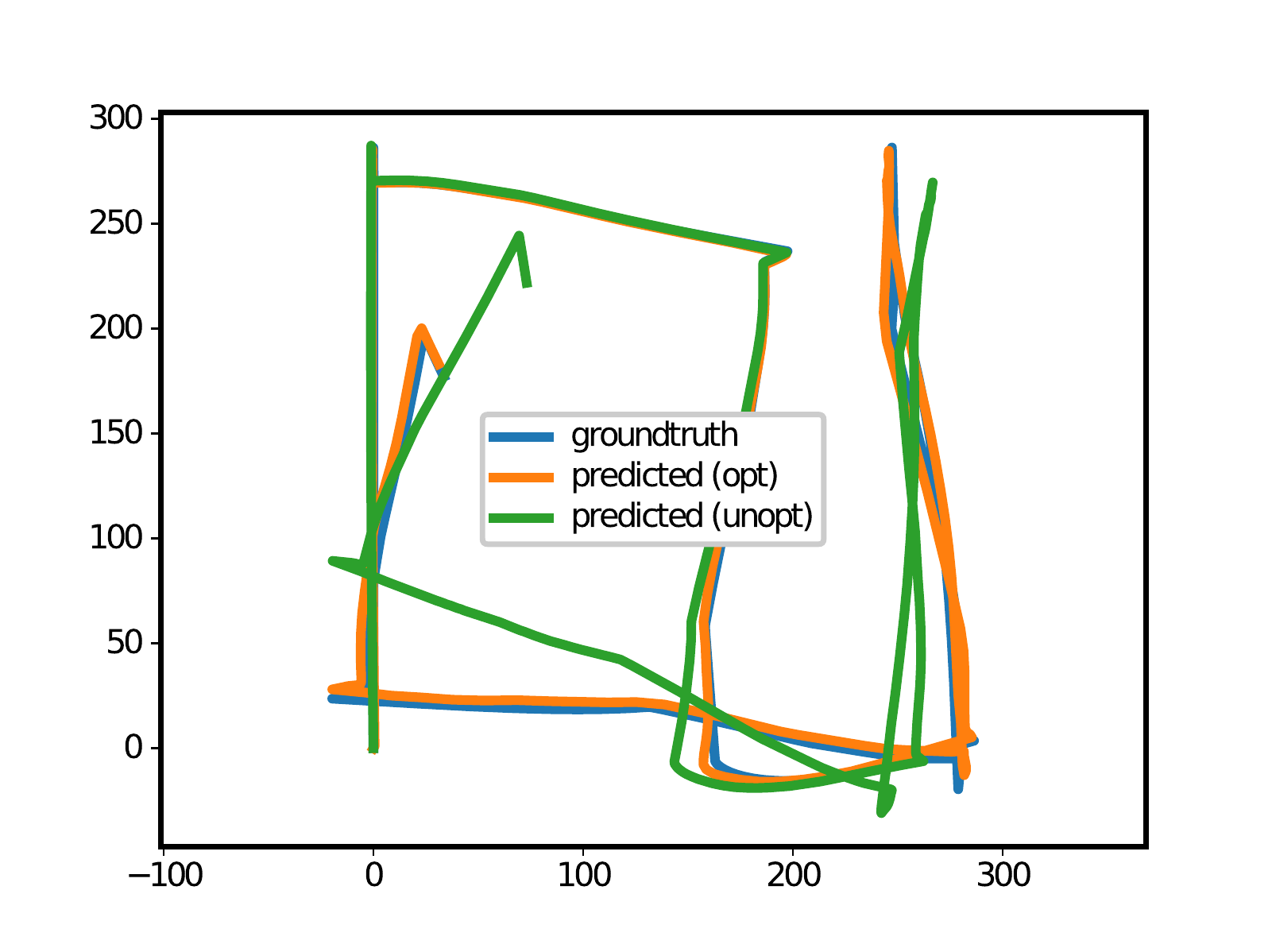} %
	\includegraphics[trim = 7mm 7mm 7mm 7mm,clip, width=0.49\linewidth]{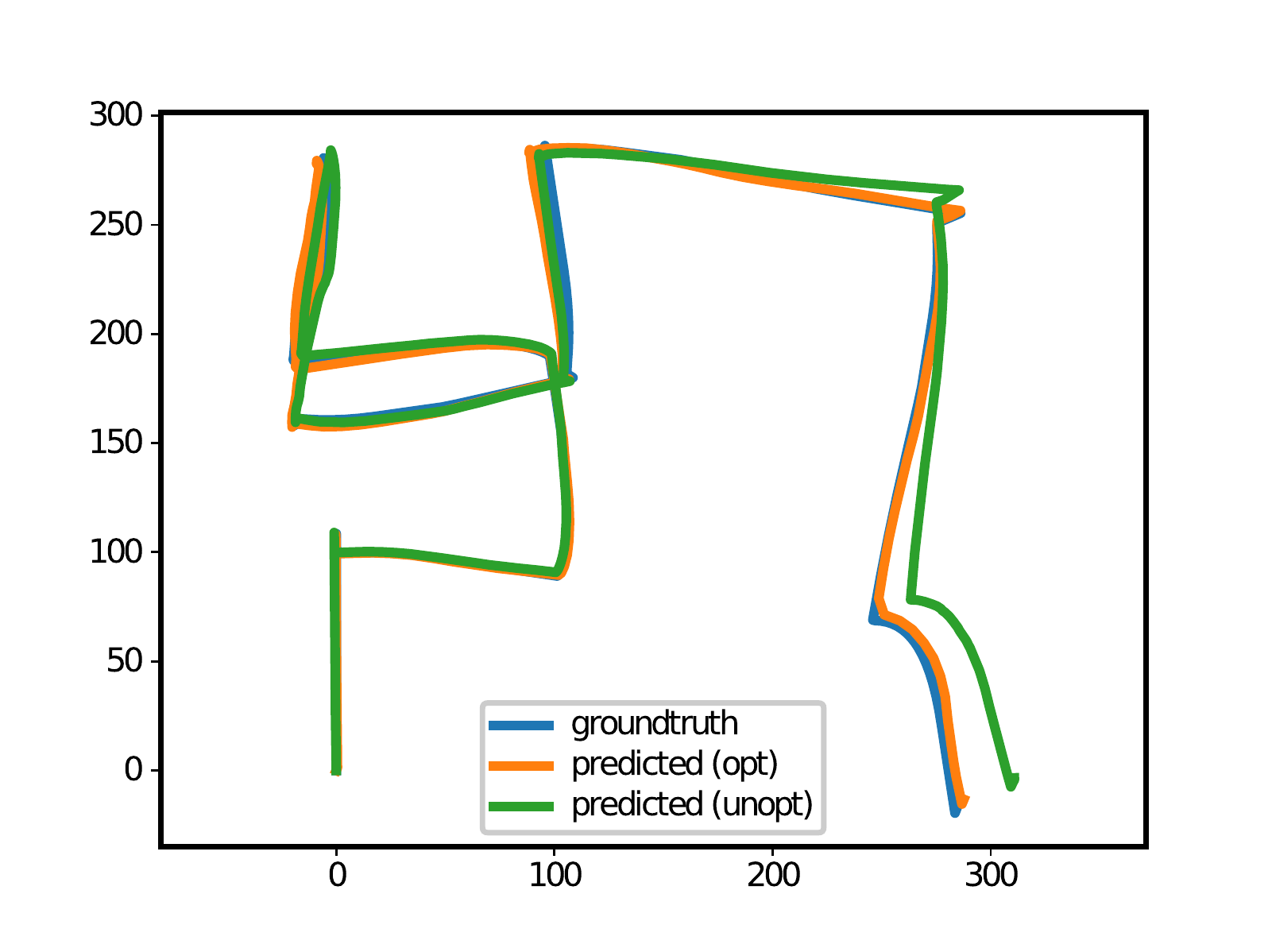} \\\vspace{0.1in}
	\centering
	\includegraphics[trim = 7mm 7mm 7mm 7mm,clip, width=0.49\linewidth]{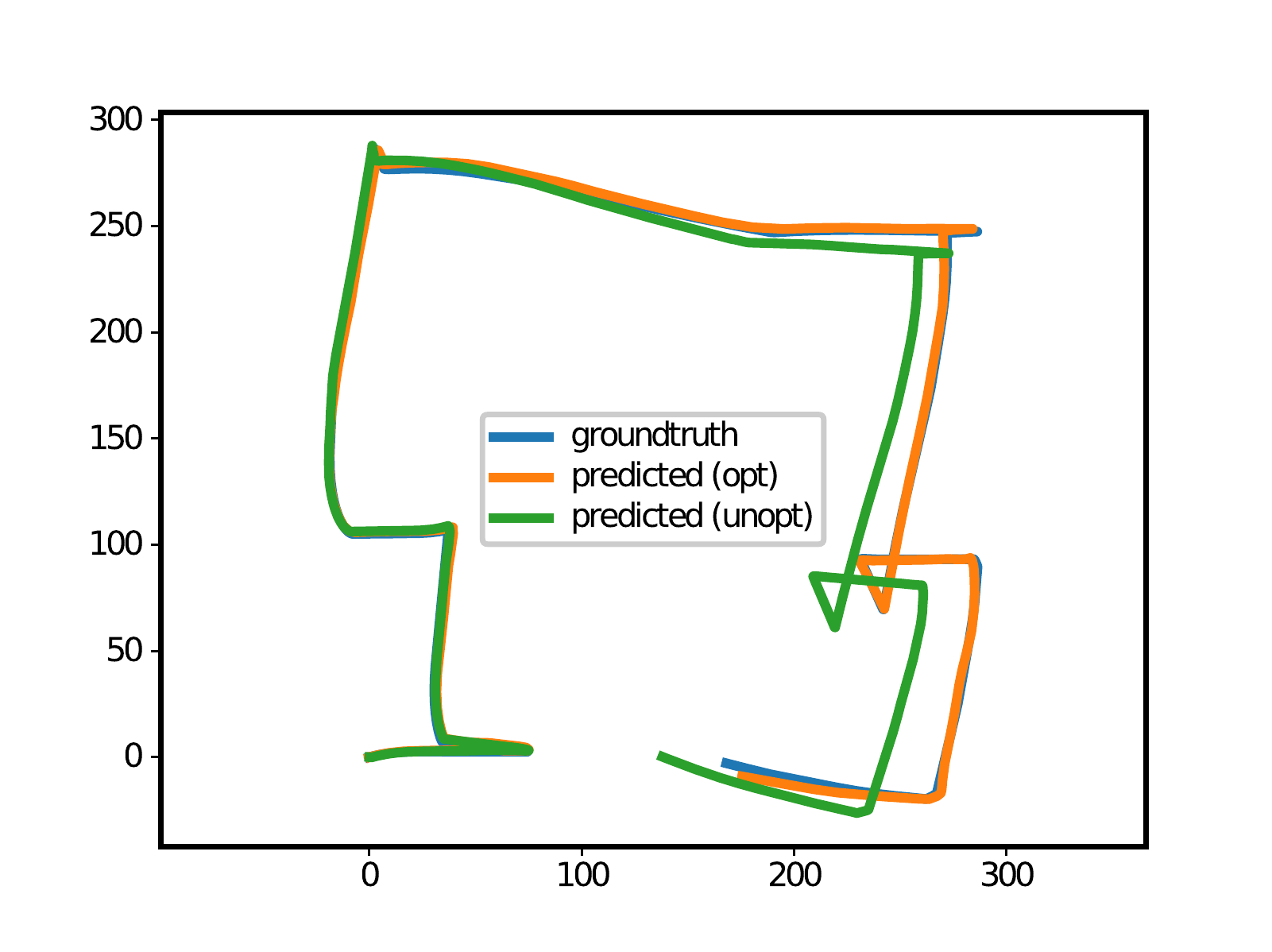} %
	\includegraphics[trim = 7mm 7mm 7mm 7mm,clip, width=0.49\linewidth]{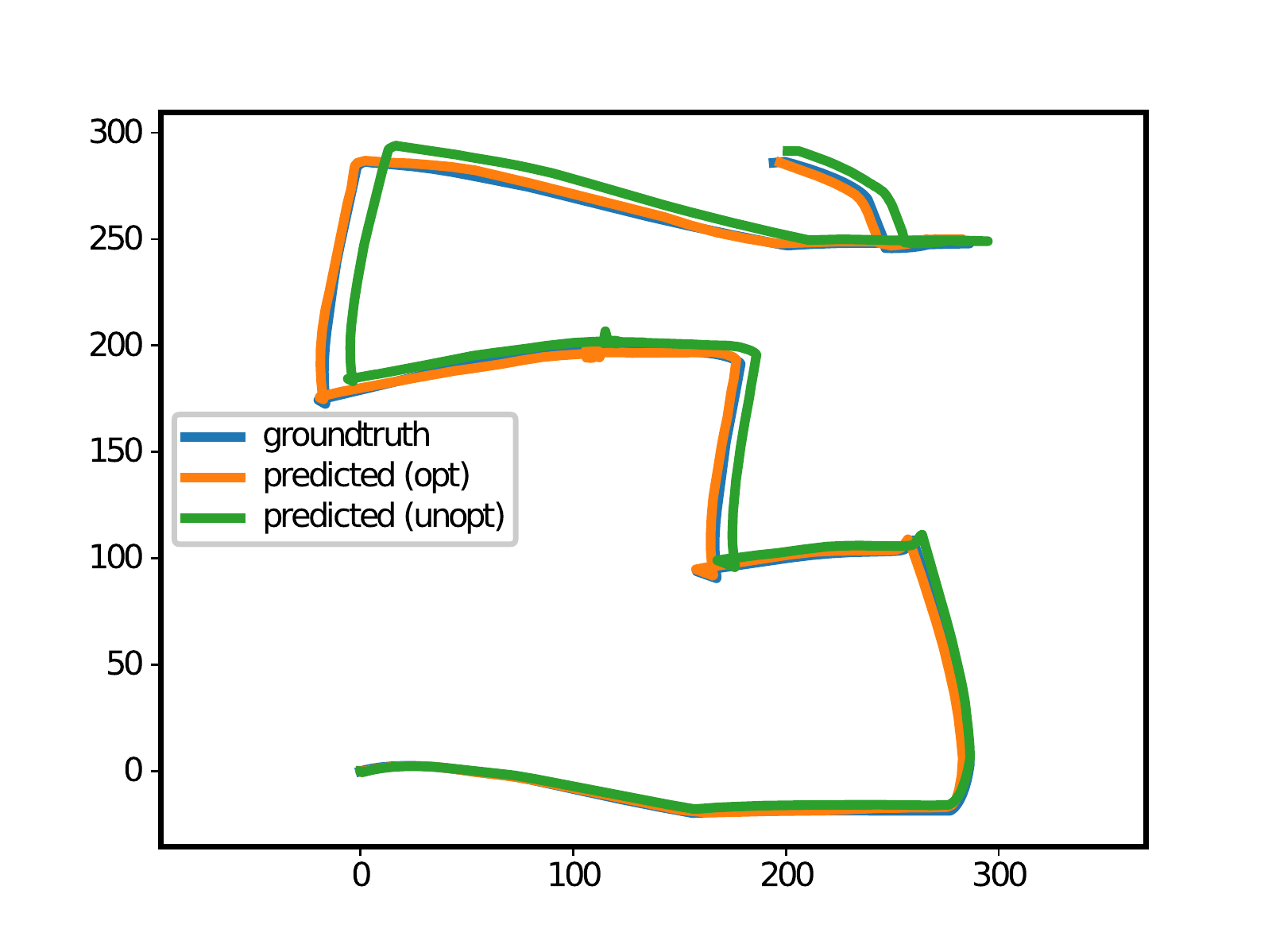}
	\vspace{0.05in}
	\caption{\label{fig:2dtraj} Images visually demonstrating the effect on pose estimates of adding the Neural Graph Optimizer module on top of the local pose estimation model in the 2D environment. We can see that the global optimization procedure greatly reduces drift. These figures were generated with the 5 iteration Neural Graph Optimization model.}
\vspace{-0.05in}
\end{figure}

\setlength{\tabcolsep}{12.4pt}
\begin{table*}
\vspace{-0.00in}
\small
  \centering
	\begin{tabular}{|c|cc|cc|}
		\hline
       \multicolumn{5}{|c|}{\bf{Results on the 3D Doom Environment }} \\ \hline
		{\bf Model} & \multicolumn{2}{|c|}{\bf Seen} & \multicolumn{2}{|c|}{\bf Unseen} \\
		& \% Err. Trans. & \% Err. Rot. & \% Err. Trans. & \% Err. Rot. \\ \hline
		Only Local Estimation                  & 1.65  & 0.117  & 1.62 & 0.122  \\ \hline
		Global Estimation - 1 Attend-Opt iteration  	       & 1.42  & 0.071  & 1.16 & 0.071 \\
		Global Estimation - 5 Attend-Opt iterations  	       & 1.25  & 0.057  & 1.04 & 0.056 \\ \hline
      		DeepVO~\cite{wang2017deepvo}                          & 1.78  & 0.079  & 2.39 & 0.091 \\ \hline
	\end{tabular}
\vspace{0.1in}
	\caption{\label{tab:doom} Results for different Neural Graph Optimizer architectures and hyperparameters, in terms of \% translation and rotation error on maps either seen or unseen during training time. We can see that the addition of the global optimization procedure reduces error significantly compared to using only the local pose model. In addition, increasing the number of attention iterations provides an increase in performance.  }
\vspace{-0.1in}
\end{table*}

\subsection{2D Environment}
For the 2D Environment, random maze designs are generated using Prim's algorithm \cite{prim1957shortest}, and the environment is created using Box2D (box2d.org). The agent projects 241 rays uniformly in front of itself with an effective field of view of 300$^{\circ}$. The observation of the agent includes the RGB values as well as the depth of the points where these rays hit a wall. An example of the 2D environment is shown in Fig.~\ref{fig:env}. Each cell in the maze has a random color. The agent can take one of three discrete actions at every time step: move-forward, turn-left, or turn-right. These actions result in translational acceleration if the action is move-forward or angular acceleration if the action is turn-left or turn-right. Data is collected by visiting four different corners on the maze using Dijkstra's algorithm \cite{dijkstra1959note}.

For this environment, the training data is generated by worker threads in parallel with the model training and each training datapoint is used only once. A test set is fixed and common for all experiments. Each epoch of training consists of $200,000$ datapoints. The error metric is Root Mean Squared Error (RMSE) in pose estimation.

To improve upon the results produced by the local pose estimation model, we train a Neural Graph Optimizer on the pose outputs of a pretrained Local Pose Estimation model. For the 2D environment, as shown in Table~\ref{tab:2dngo}, we observed over 80\% improvement in the correction of drift compared to using only the local pose estimation model, as measured by the root mean squared error loss. We can see that increasing the number of iterations (applying the attention operator and then the temporal aggregation operator) improved results from 1 to 5 iterations. 
We show some sample trajectories in Fig.\ \ref{fig:2dtraj} before and after the Neural Graph Optimizer procedure.

\subsection{3D Environment}
For the 3D Environment, random maze designs are generated using the Kruskal's algorithm \cite{kruskal1956shortest}, and the environment is created using the ViZDoom API \cite{kempka2016vizdoom}. The agent observes the environment in a first-person view with a field-of-view of 108$^{\circ}$. An example of the 3D environment design 
is shown in Fig.\ \ref{fig:env}. Similr to the 2D environment, the pose predictions are 3-dimensional tuples (x, y, angle) and the agent can take one of three discrete actions at every time step: move-forward, turn-left, or turn-right, which results in translational or angular acceleration. For collecting data in this environment, a navigation network \cite{lample2017playing} is trained to maximize the distance travelled by the agent using the Asynchronous Advantage Actor-Critic algorithm \cite{mnih2016asynchronous}. The data is collected by using the policy learned by the navigation network.

Like the 2D environment, the training data is generated by worker threads in parallel with the model training, and each training datapoint is used only once. We additionally sample two test sets, one containing 39 trajectories sampled from maze geometries that were seen during training and one containing 39 trajectories sampled from novel maze geometries that the agent had not encountered during training.

\subsubsection{Results}
\begin{figure}[t]
\vspace{-0.1in}
        \centering
        \includegraphics[width=0.5\linewidth]{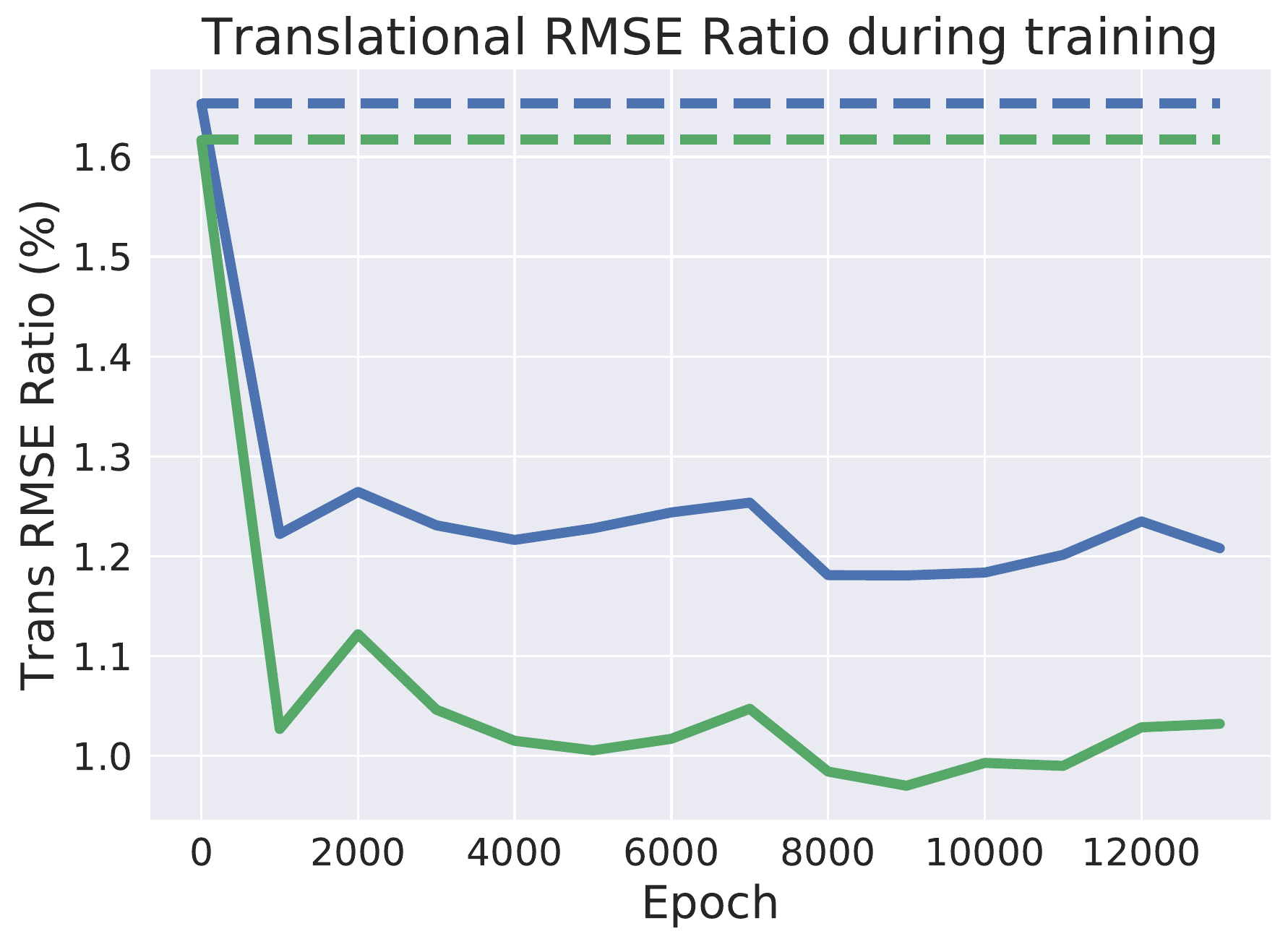}%
        \includegraphics[width=0.5\linewidth]{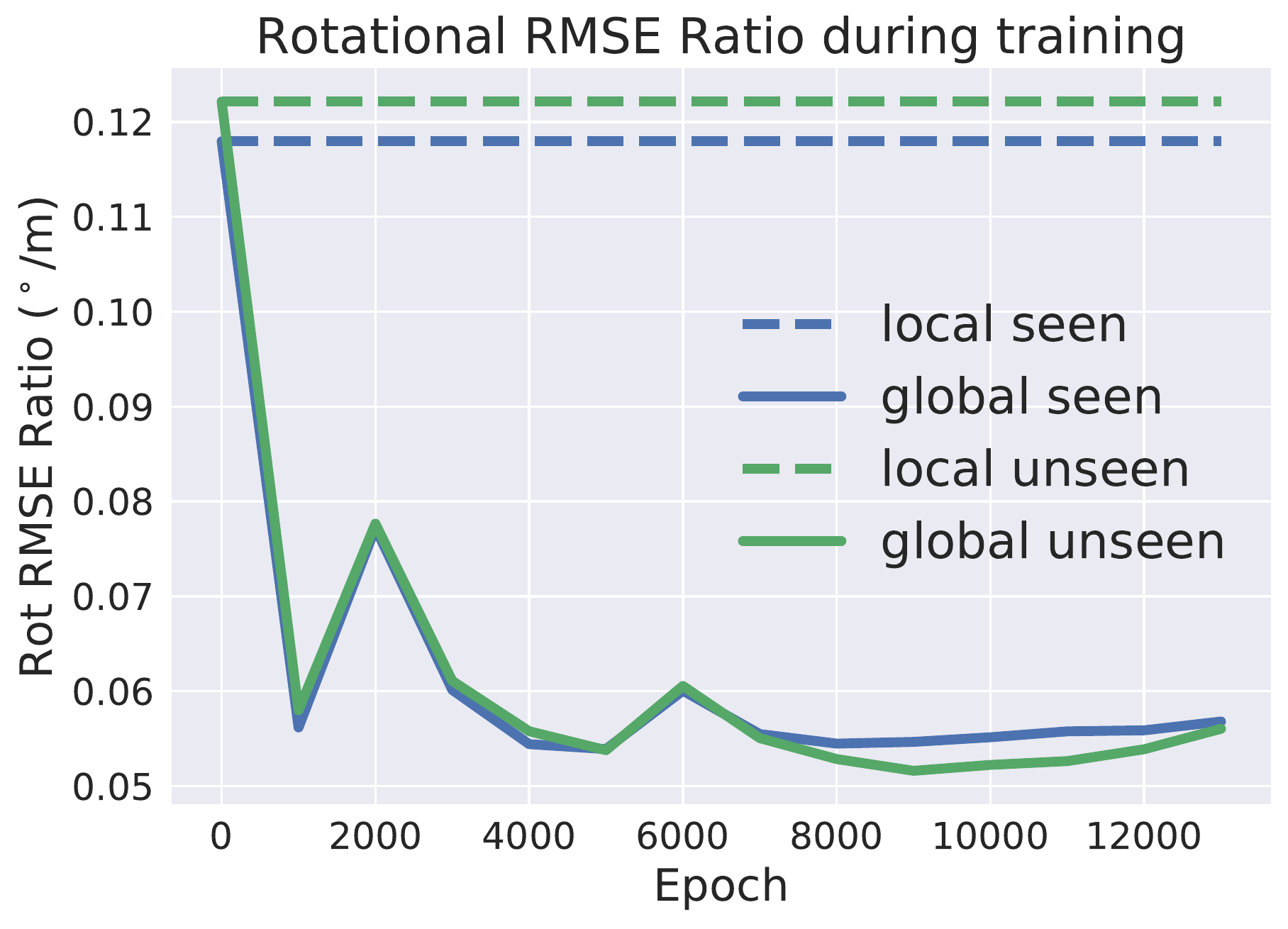}
        \caption{\label{fig:doomcurves} Training curves for Doom over $13,000$ updates for the~5 iteration Attend-Opt model. We show the performance on both seen and unseen test sets as training progress. The dotted line represents the estimate provided by using only the local model. We can see there is a large reduction in error when making use of the global optimizer.}
\vspace{-0.05in}
\end{figure}

Results are shown in Table~\ref{tab:doom}. Here we report~\%~Error in Translation and Rotation for seen/unseen mazes, where the accumulated drift error is divided by the entire distance traveled in each trajectory. Observe that the local model is significantly improved by using global optimization and performance of the global model improves as we increase the number of Attend-Opt iterations from 1 to 5. The global model outperforms the DeepVO~\cite{wang2017deepvo} baseline on both the test sets. Additionally, we can clearly see that the model itself does not overfit to the training environments it experienced, and gets similar or even lower error on unseen test mazes. Learning curves are shown in Fig~\ref{fig:doomcurves}. We can see that performance plateaus decrease significantly early on and then progress is much slower after around 2000 updates.

The baseline DeepVO~\cite{wang2017deepvo} is one of the state-of-the-art methods using deep neural nets for monocular visual odometry. It stacks 2 consecutive frames and passes them through 9 convolutional layers followed by 2 LSTM layers to estimate the pose changes. As compared to the proposed Local Pose Estimation model which observes only the last 2 frames at the time, the DeepVO model can potentially utilize information from all the prior frames using the LSTM layer. However, the DeepVO model does not correct its previous predictions as it observes new information. The Neural Graph Optimizer has the ability to correct its predictions using the Attention operation and consequently leads to improved performance.

\subsubsection{Analysis}
\begin{figure}[t]
  \centering
  \minipage{0.5\linewidth}
    \centering
    \includegraphics[width=\linewidth,height=\textheight,keepaspectratio]{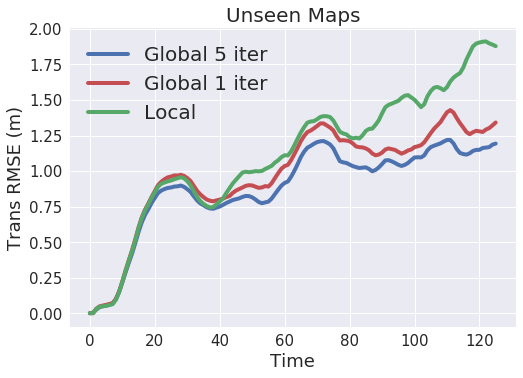}
  \endminipage \hfill
  \minipage{0.5\linewidth}
    \centering
    \includegraphics[width=\linewidth,height=\textheight,keepaspectratio]{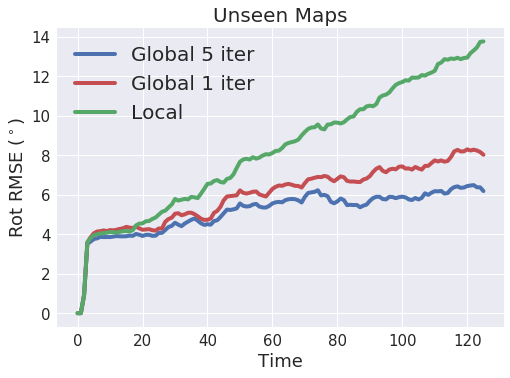}
  \endminipage
        \caption{\small{Translational (Left) and Rotational (Right) RMSE as a function of number of images in the trajectory in {\bf unseen mazes}.}}
  \label{fig:plot_unseen}
\end{figure}

\begin{figure}[t]
  \centering
  \minipage{0.5\linewidth}
    \centering
    \includegraphics[width=\linewidth,height=\textheight,keepaspectratio]{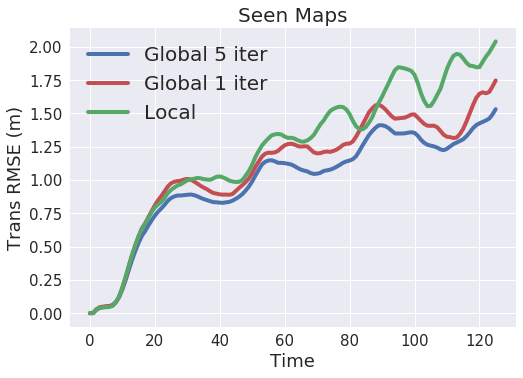}
  \endminipage \hfill
  \minipage{0.5\linewidth}
    \centering
    \includegraphics[width=\linewidth,height=\textheight,keepaspectratio]{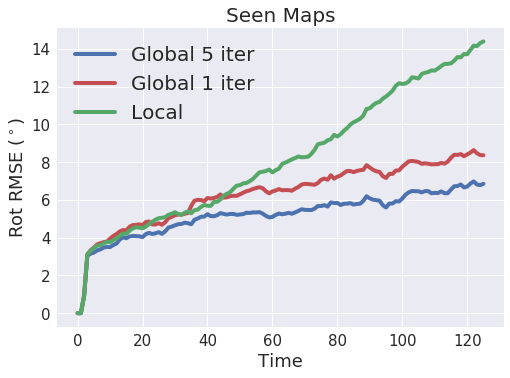}
  \endminipage
        \caption{\small{Translational (Left) and Rotational (Right) RMSE as a function of number of images in the trajectory in {\bf seen mazes}.}}
  \label{fig:plot_seen}
\end{figure}

\begin{figure}[t!]
  \centering
  \minipage{0.5\linewidth}
    \centering
    \includegraphics[width=\linewidth,height=\textheight,keepaspectratio]{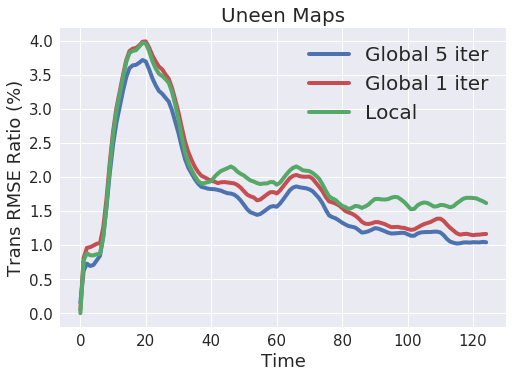}
  \endminipage \hfill
  \minipage{0.5\linewidth}
    \centering
    \includegraphics[width=\linewidth,height=\textheight,keepaspectratio]{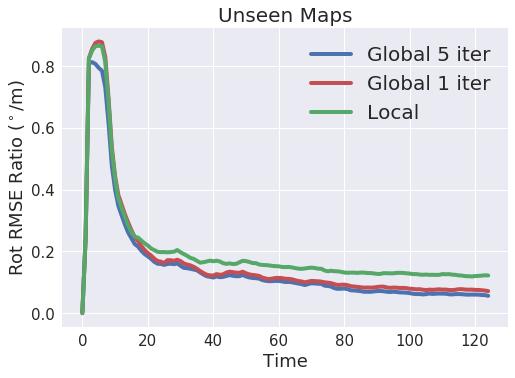}
  \endminipage
	\caption{\small{Ratio of the Translational (Left) and Rotational (Right) RMSE to the distance travelled as a function of number of images in the trajectory in {\bf unseeen mazes}.}}
  \label{fig:plot_ratio_unseen}
\end{figure}

\begin{figure}
  \centering
  \minipage{0.5\linewidth}
    \centering
    \includegraphics[width=\linewidth,height=\textheight,keepaspectratio]{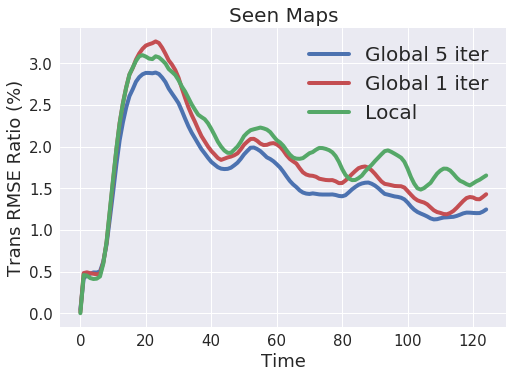}
  \endminipage \hfill
  \minipage{0.5\linewidth}
    \centering
    \includegraphics[width=\linewidth,height=\textheight,keepaspectratio]{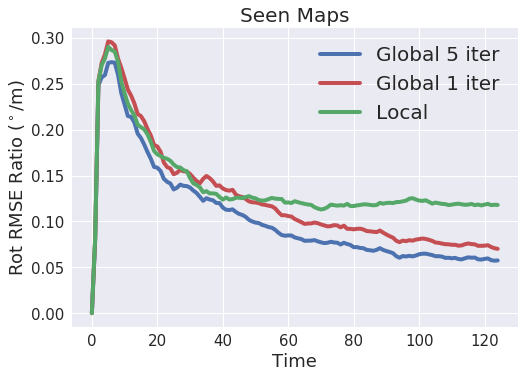}
  \endminipage
	\caption{\small{Ratio of the Translational (Left) and Rotational (Right) RMSE to the distance travelled as a function of number of images in the trajectory in {\bf seen mazes}.}}
  \label{fig:plot_ratio_seen}
\end{figure}

We next plot the total rotational and translational errors as a function of number of steps in the trajectory in Figures~\ref{fig:plot_unseen} (for unseen mazes) and ~\ref{fig:plot_seen} (for seen mazes). The global model reduces the slope of the rate of increase of both translational and rotation errors as compared to the local estimates. Figures~\ref{fig:plot_ratio_unseen} and~\ref{fig:plot_ratio_seen} display the ratio of the translational (left) and rotational (right) drift error over distance traveled. We can see from these plots that the trend is negative, meaning that drift accumulates much slower than the distance being traveled. This indicates that the model is likely to generalize well to arbitrarily long trajectories. Additionally, in all plots, we can see a clear ordering of the performance of the models, where the local model performs worst, one iteration of Attend-Opt increases model performance significantly, and increasing the number of Attend-Opt iterations to 5 further increases model performance. 
 
\begin{figure}
	\centering
	\includegraphics[trim = 7mm 7mm 7mm 7mm,clip, width=0.49\linewidth]{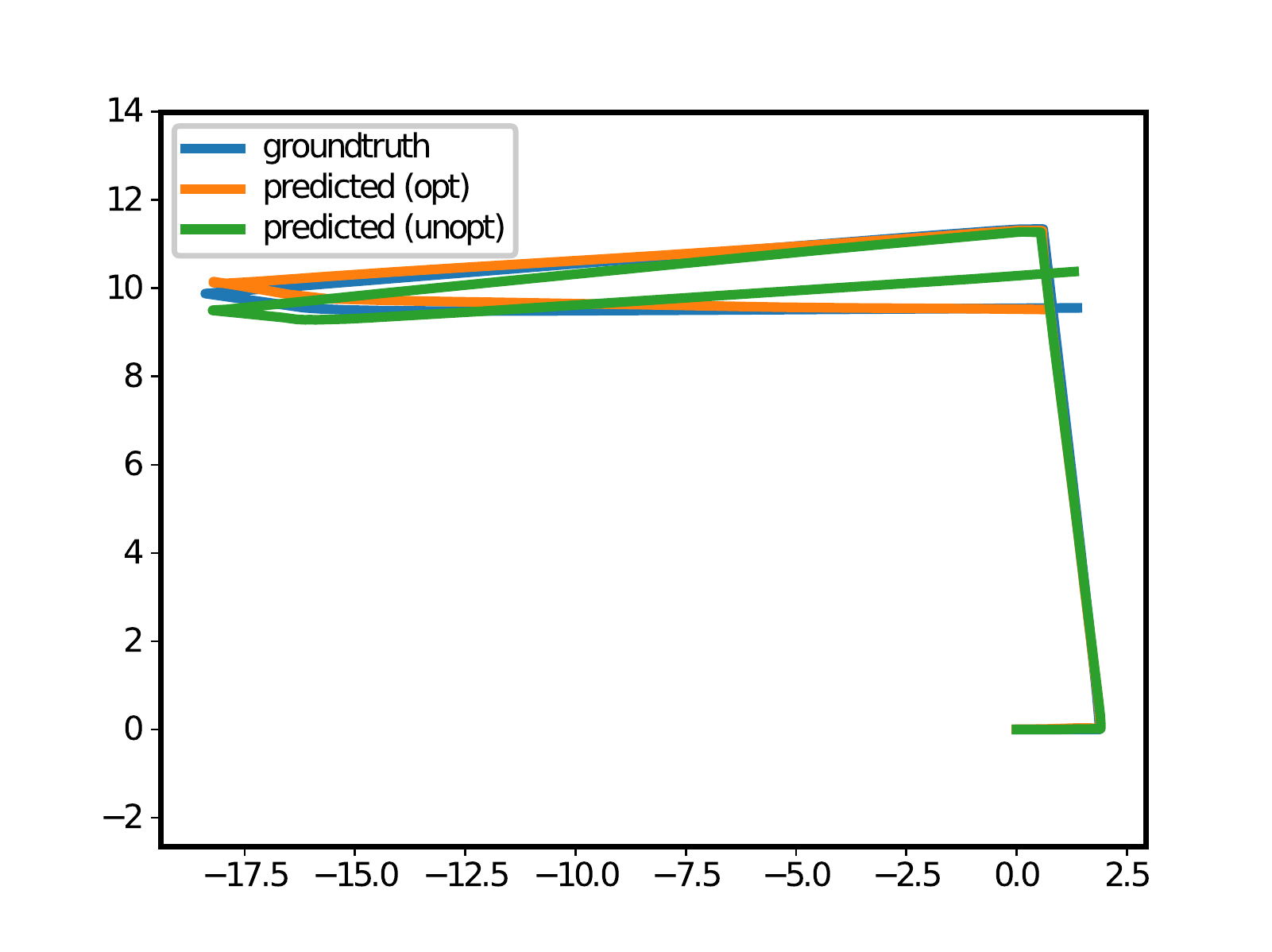} %
	\includegraphics[trim = 7mm 7mm 7mm 7mm,clip, width=0.49\linewidth]{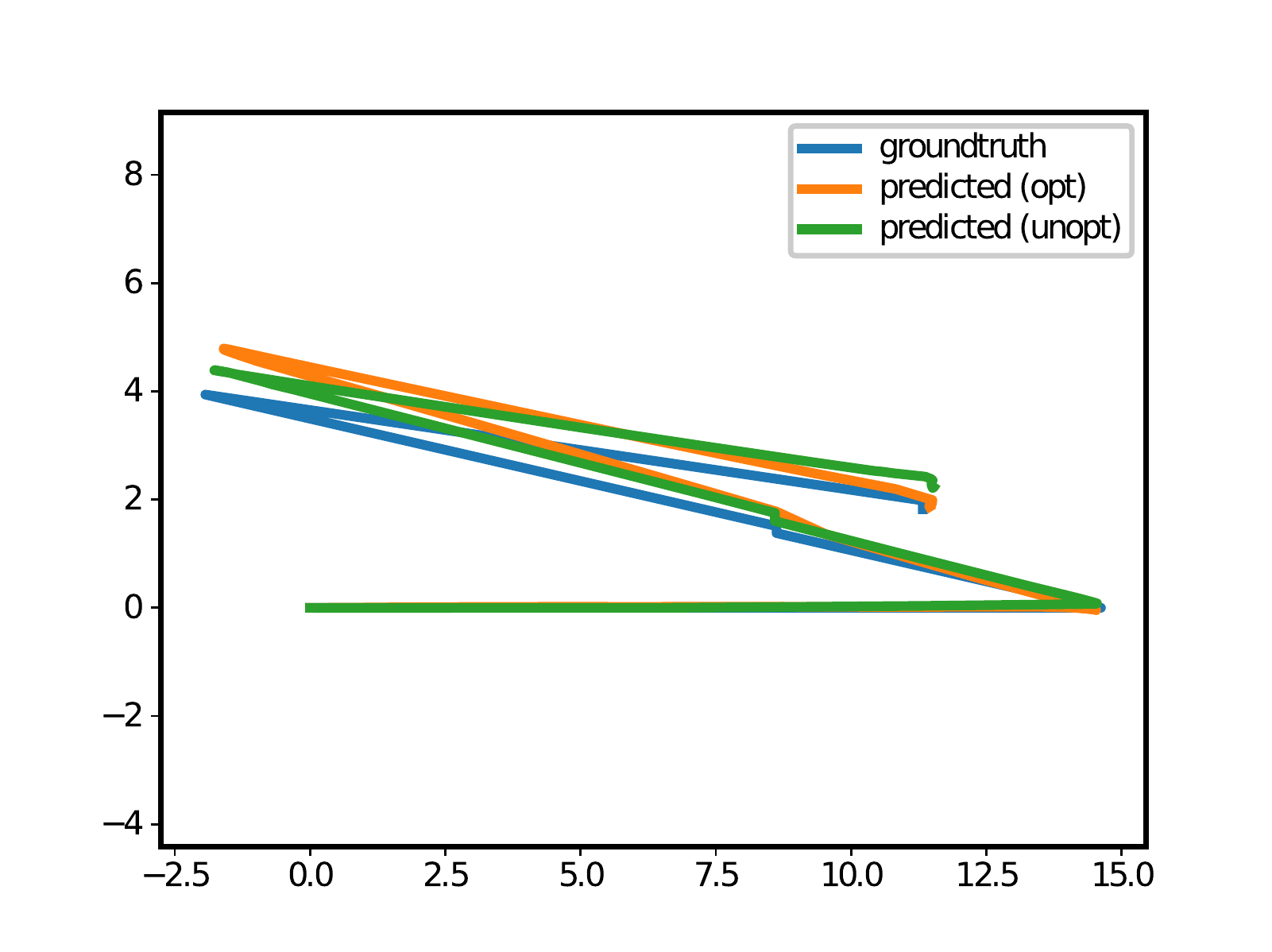} \\\vspace{0.1in}
	\centering

	\includegraphics[trim = 7mm 7mm 7mm 7mm,clip, width=0.49\linewidth]{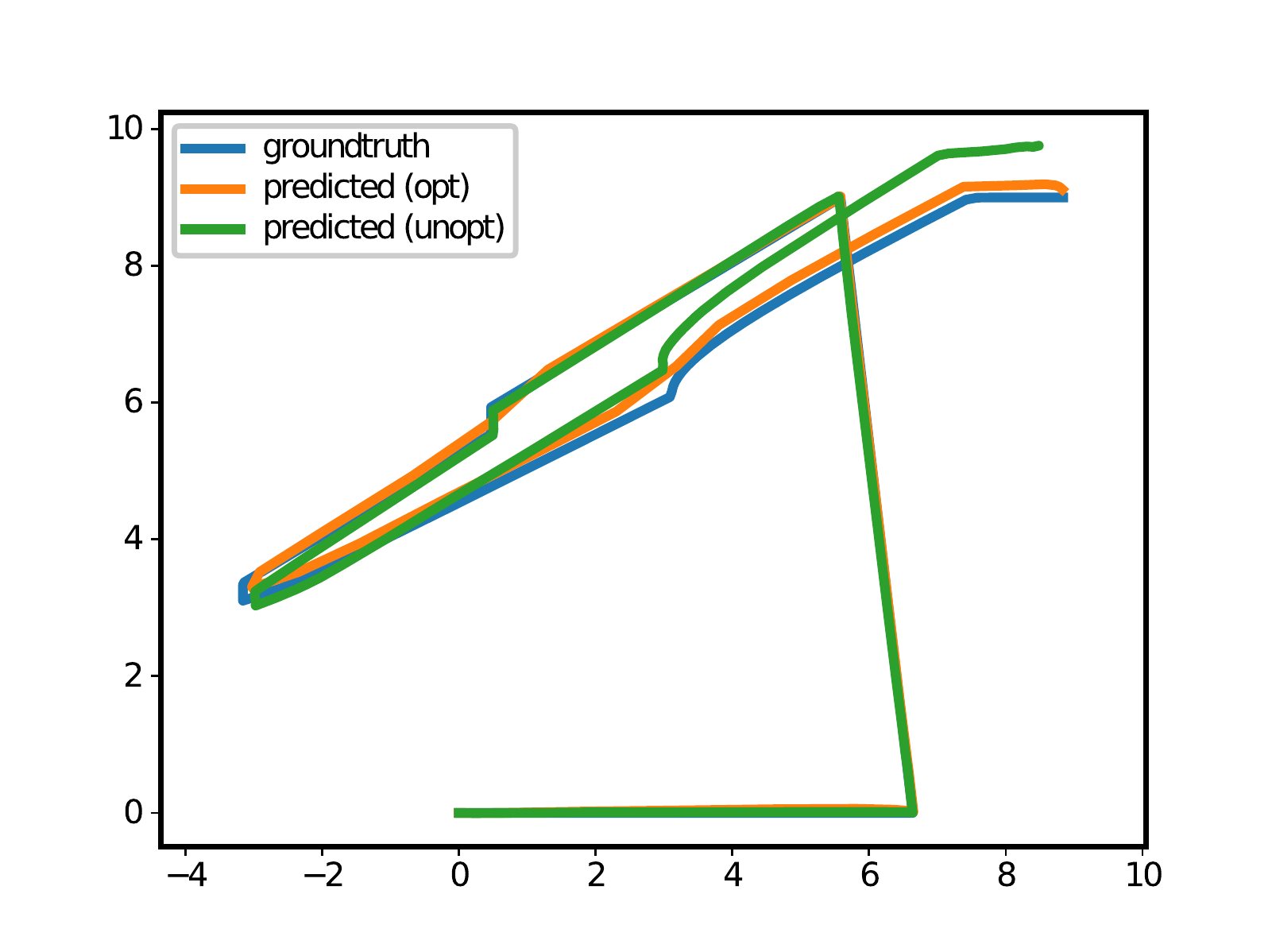} %
	\includegraphics[trim = 7mm 7mm 7mm 7mm,clip, width=0.49\linewidth]{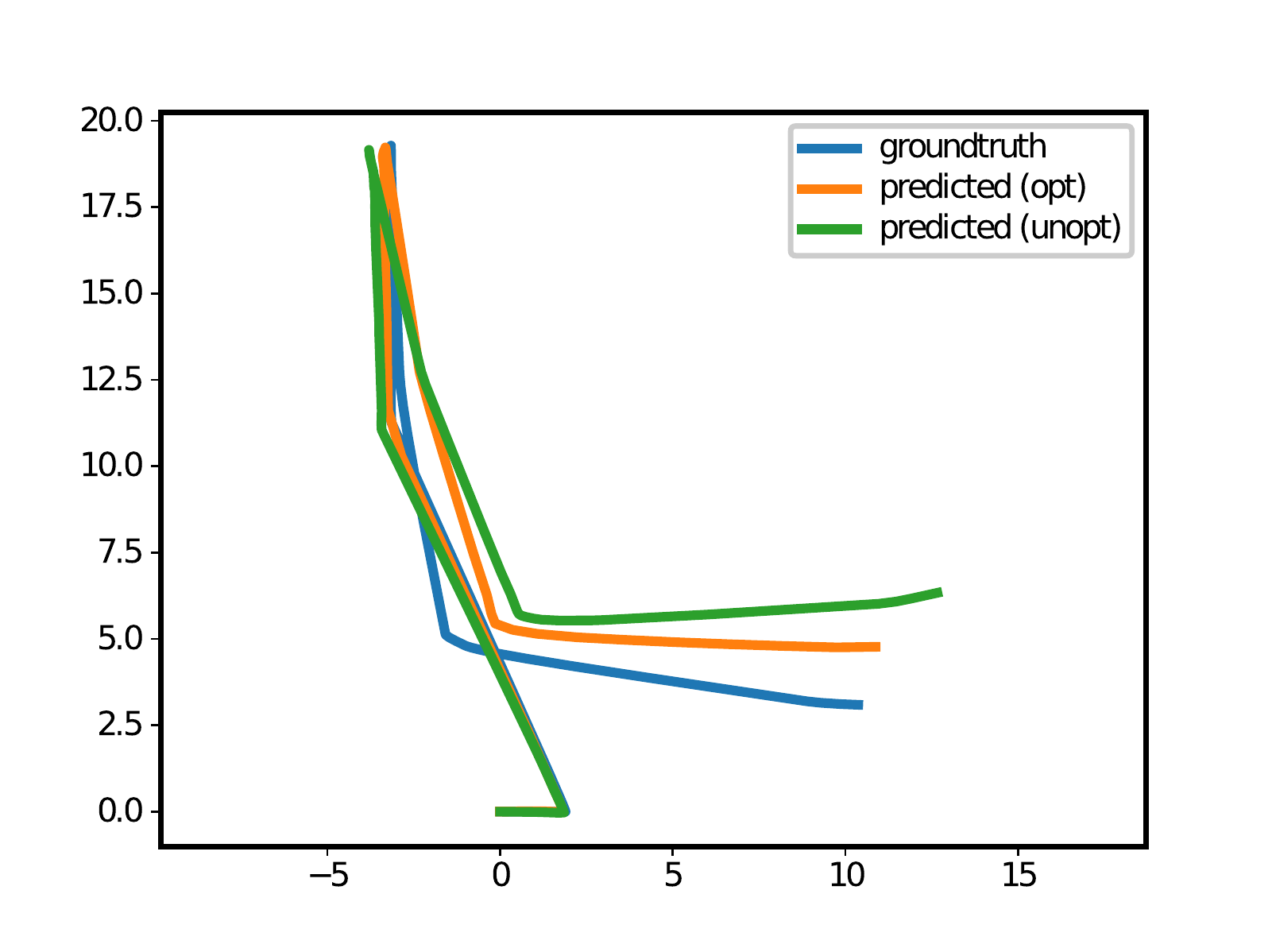}
	\vspace{0.05in}
	\caption{\label{fig:3dtraj} Images visually demonstrating the effect on pose estimates of adding the Neural Graph Optimizer module on top of the local pose estimation model in the 3D environment. We can see that the global optimization procedure greatly reduces drift. These figures were generated with the 5 iteration Neural Graph Optimization model. The agent always starts at the origin (0, 0).}
\end{figure}

The plots in Figures~\ref{fig:plot_unseen} and ~\ref{fig:plot_seen} as well as the numbers in Table~\ref{tab:doom} show that the improvement in rotational errors due to the neural optimization is higher than the improvement in translation errors. Fig~\ref{fig:3dtraj} shows sample trajectories with estimates of both global and local pose estimates. As seen in the figure, the neural graph optimizer considerably improves the rotation estimates, consequently leading to significant improvements in the drift reduction.

\section{Conclusion}
  
In this paper, we designed a novel attention-based architecture to perform an end-to-end trainable global pose estimation. Compared to the previous work on using deep networks to do pose estimation, our method uses an attention operation to re-estimate its trajectory at each time step and therefore enables iterative refinement of the quality of its estimates as more data is available.
We demonstrate the benefit of the model on two simulators, the first is a top-down 2D maze world and the second is a 3D random maze environment running the Doom engine.
Our results show that our method has an increased performance compared to models which used only temporally local information.

The proposed method can  be further extended to a complete end-to-end graph-based SLAM system by adding a relocalization module which uses pose features to relocalize in a known map~\cite{chaplot2018active}. It can also be extended to an Active SLAM system where the agent also decides the actions, in order to map the environment as fast as possible.

\subsubsection*{Acknowledgments}

We thank Tim Barfoot and Russ Webb for helpful comments and discussions. We would also like to thank Barry Theobald and Megan Maher for helpful feedback on the manuscript.

{\small
\bibliographystyle{ieee}

}

\end{document}